\documentclass[acmlarge]{acmart}
\usepackage{subfigure}
\usepackage{multirow}
\usepackage{romannum}

\AtBeginDocument{%
  }

\setcopyright{acmlicensed}
\copyrightyear{2018}
\acmYear{2018}
\acmDOI{XXXXXXX.XXXXXXX}

\acmJournal{POMACS}
\acmVolume{37}
\acmNumber{4}
\acmArticle{111}
\acmMonth{8}

\begin{document}

\title{{MISO:} Monitoring Inactivity of Single Older Adults at Home {using RGB-D Technology}}

\author{Longfei Chen}
\email{longfei.chen@ed.ac.uk}
\orcid{0000-0002-3935-802}
\author{Robert B. Fisher}
\email{rbf@inf.ed.ac.uk}
\orcid{0000-0001-6860-9371}
\affiliation{%
  \institution{The University of Edinburgh}
  \streetaddress{47 Potterrow}
  \city{Edinburgh}
  \state{Scotland}
  \country{UK}
  \postcode{EH8 9BT}
}

\renewcommand{\shortauthors}{Chen et al.}

\begin{abstract}
    A new application for real-time monitoring of the lack of movement in older adults' own homes is proposed, aiming to support people's lives and independence in their later years.  A lightweight camera monitoring system, based on an RGB-D camera and a compact computer processor, was developed and piloted in community homes to observe the daily behavior of older adults.  Instances of body inactivity were detected in everyday scenarios anonymously and unobtrusively. These events can be explained at a higher level, such as a loss of consciousness or physiological deterioration. The accuracy of the inactivity monitoring system is assessed, and statistics of inactivity events related to the daily behavior of older adults are provided. {The results demonstrate that our method achieves high accuracy in inactivity detection across various environments and camera views. It outperforms existing state-of-the-art vision-based models in challenging conditions like dim room lighting and TV flickering. However, the proposed method does require some ambient light to function effectively.}
\end{abstract}

\ccsdesc[500]{Applied computing~Health informatics}

\keywords{Anonymously monitoring, Inactivity, Older adults, Computer Vision, Aging in place}

\received{10 November 2023}
\received[revised]{12 March 2009}
\received[accepted]{5 June 2009}

\maketitle

\section{Introduction}
\label{sec:introduction}

The proportion of older adults living alone is increasing globally as the aging population grows \cite{b10, b11}. 
According to the Office for National Statistics in 2019, more than 3 million people over the age of 70 lived alone in the UK \cite{b2}.
Even if they are living alone, more than 90\% of older adults express a strong desire to remain independent -- 
they predominantly prefer to continue living in their own homes rather than relocating to nursing homes or other care facilities \cite{b3, b6}. 
However, older adults living alone face more difficulties in daily life and have higher medical needs \cite{b10}, \cite{b5}. 
Meanwhile, new or worsening symptoms related to chronic health problems or sensory impairments may not be noticed \cite{b3}.

\begin{figure}[t]
    \centering
	\includegraphics[width=.7\columnwidth]{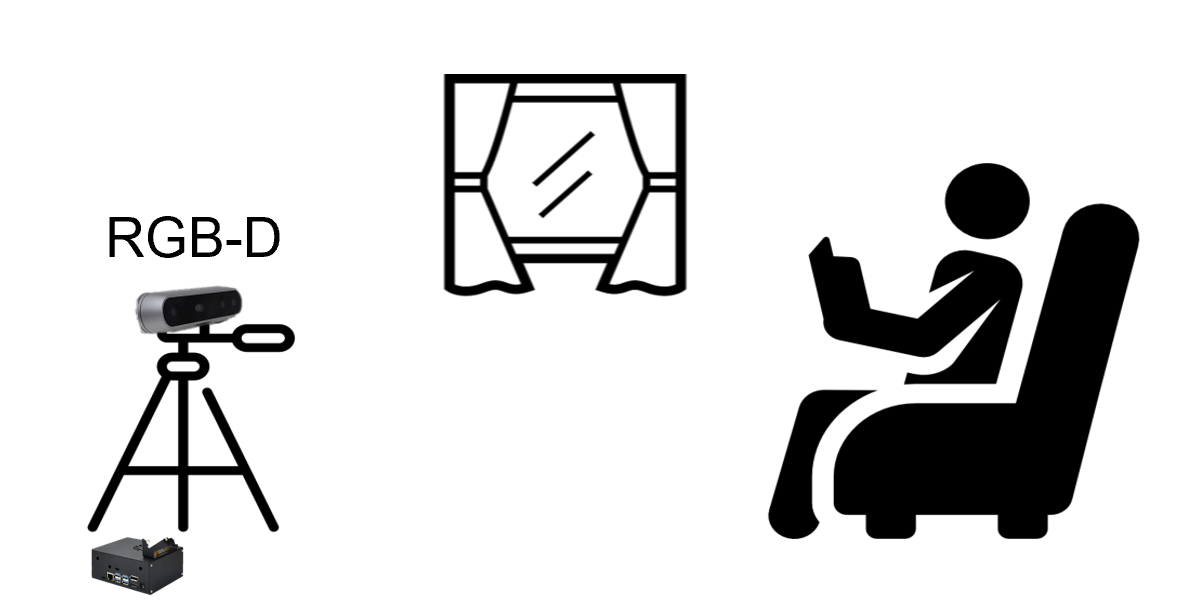}
    \caption{Real-time system for monitoring older adults in common home scenarios (inactivity monitoring of routine seating areas).
Anonymous monitoring uses a compact system consisting of an RGB-D camera and a small computer processor. Anonymity is maintained by discarding captured images after processing.}
    \label{fig.1}
\end{figure}

Mobility problems, which encompass a spectrum of limitations including complete immobility (immobility syndrome) and reduced ability to move, can arise when someone experiences a prolonged period of reduced movement. These limitations are particularly prevalent in older adults and can contribute to functional decline, an increased risk of needing long-term care after hospitalization, and various medical complications. Examples include deep vein thrombosis, urinary incontinence, pressure sores, joint contractures, cardiac deconditioning, and muscle weakness \cite{immor}. {As these complications often develop gradually, our focus lies on monitoring and analyzing behavior in older adults with long-term mobility limitations.}

Smart technologies and AI solutions are accelerating the pace of change in healthcare in many areas \cite{b12}, \cite{b14}, \cite{b15}, \cite{b16}.
Unobtrusive sensors can work 24/7 in the long term, providing faster, more accurate analysis in a user-focused manner.
In recent years, many studies have been conducted to monitor older adults in their homes using camera sensors.
Stone \emph{et al.} \cite{b14} detect falls using the person's vertical state and motion features. 
Using a depth imagery sensor (Kinect) can largely eliminate the interference of lighting and shadows for visual perception. 
A patient monitoring system is used to lessen the workload of the nurses: 
two RGB cameras were placed for monitoring patients, one for bed view and the other for room view, 
to detect bed occupancy, self-extubation, and falls \cite{b15}.
The above systems have achieved good accuracy in detecting events, 
however, one of the major concerns with camera-based systems is  \textit{privacy} \cite{b17}. 
This issue must be addressed when convincing people to use such a camera-based system.
Other challenges of real-world home camera monitoring include effectiveness and adaptation. 
That means accurate detection of anomalous human events in backgrounds for different scenarios, moving objects and pets in the house, 
and changing lighting conditions, such as sunlight, low light, and TV lighting.

{In this study, a new application of real-time monitoring of older adult people in home scenarios is presented. 
Specifically, the application detects the prolonged inactivity of a person sitting in a standard home location (e.g., on a favorite chair).  
A compact system consisting of an RGB-D camera and a small computer is deployed, which monitors the person at a specific location.
The system is camera-based, but completely anonymous, with no internet connection, no image/video is saved, and only inactivity statistics text logs are kept. 
The abnormal inactivity events can be related to medical conditions, such as lost consciousness, or long-term decreased physical ability. 
An example is shown in Fig. \ref{fig.1}.}

{The application was designed for homes where a single aging adult lives, so this is the test scenario that is evaluated. 
The results show that the system has good accuracy, sensitivity, and robustness in different environmental settings, 
enabling long-term anonymous monitoring of older adults in their own homes.}

{
This paper introduces a new visual monitoring system that has the following advantages:}


{(1) A new camera system automatically tracks older adults' inactivity at home, even in difficult situations like dim light, pets around, or TV flickering. 
It works in real-time and keeps identities anonymous. 
This system is more accurate than other state-of-the-art vision algorithms for inactivity detection.}

{(2) The system was piloted in community homes to unobtrusively detect instances of body inactivity in everyday scenarios, revealing the behavior patterns of older adults, such as their daily routines and motion habits.}

{(3) This affordable, reliable, and zero-interacting system could be used to monitor older adults suffering from frailty and other long-term physical deterioration, as well as detect critical situations, making it a useful tool for everyday health monitoring.}


\section{Related Work}\label{sec2}

{Many sensors have been investigated for detecting human behavior in indoor scenarios \cite{b19}. Individual inactivity is closely associated with hospitalization in older adults \cite{immo}. However, few works specifically focus on detecting the inactivity of the human body.} 

{Wearable sensors offer high-accuracy human motion detection. 
For example, a belt-worn kinematic sensor demonstrated 100\% recall in detecting human motionless events, such as several seconds of walking, standing, sitting, or lying, in a lab environment \cite{belt}. 
However, wearable sensors can be intrusive and may require users to wear them continuously, which might not always be acceptable for older adults.
Ambient sensors are non-intrusive and capable of detecting various indoor human movements, including whole-body motions, limb movements and chest breathing movements, even in complete darkness. 
For instance, Wi-Fi-based human motion sensing \cite{wifi1} can accurately recognize five typical human activities with a 96.6\% accuracy; and subtle motions, such as simulated hand tremors, with 95.7\% accuracy \cite{wifi2}.
Passive Infrared (PIR) sensors \cite{PIR} have achieved an accuracy of 93\% in predicting human relative locations, including stationary individuals. Radar-based sensors \cite{radar} have achieved a classification accuracy of over 95\% for four basic types of human motion. However, ambient sensors can be triggered by household pets, leading to an increase in data noise and a decrease in the overall predictability of human mobility \cite{pets}.}
{Camera-based systems are widely used due to their cost-effectiveness and non-redundant nature in modern buildings. 
Cameras are less intrusive for long-term monitoring compared to wearable sensors and provide multifunctional and semantically explainable capabilities.
{They can distinguish between human and non-human movement and identify which body part is moving, and motion scale/speed; they can also efficiently identify and track humans, pets, and various objects. Moreover, the same camera-based system can be used to process different tasks, where ambient motion sensors can only detect some unspecified motion in the environment.}
 For instance, Kinect-based body motion signals have shown moderate to excellent accuracy, with root mean square errors (RMSE) ranging from 20 mm to 89 mm \cite{cvacc1}. In video-based pose estimation, mean absolute errors for gait analysis are as low as 0.02 seconds for temporal gait parameters and 0.04 meters for step lengths compared to motion capture technologies \cite{cvacc2}.}

{For human inactivity detection, accurately detecting the person is typically the initial step in monitoring behaviors. Xia \emph{et al.} \cite{b23} proposed a model-based approach for indoor person detection using a Kinect sensor to capture a side view of the person. Initially, all 2D circular shapes are localized as head candidates, and then these candidate regions are fitted to a learned 3D human head shape. Most deep learning-based detectors are designed for oblique or front views, but Cho \emph{et al.} \cite{b25} trained convolutional neural networks to segment heads using top-view depth data.
A side view of the person can be more intrusive, as individuals are aware of the camera's presence. Depth-imaging-based methods are recognized for their superior performance and robustness in detecting humans across various poses, rotations, and lighting conditions. However, they often come with higher computational costs, which can be a limitation in applications requiring real-time processing, such as healthcare applications.}

To monitor a person's inactivity while sitting in a preferred location, camera-based motion detection methods can be applied. 
There is a range of motion detection algorithms designed to address various real-world challenges \cite{b29}. One fundamental method is background subtraction \cite{b31}, which creates a background model and identifies foreground objects by comparing the current frame to this model. Another technique, frame differencing \cite{b40}, involves comparing the current frame with a reference frame to track the number of differing pixels. However, these methods are sensitive to noise and environmental changes, such as variations in lighting, the presence of shadows, and moving objects. Parametric-based methods, like the Gaussian Mixture model \cite{b34}, tend to be more robust in the face of noise and artifacts \cite{b32}. Non-parametric methods \cite{b33}, fitting a smooth probability density function to pixel values over a temporal window, considering both self-similarity and similarity to neighboring pixels. This enhances robustness against camera jitter or minor background movements \cite{b29}. When compared to other traditional motion detection methods, non-parametric techniques have shown superior performance in eliminating minor background movements \cite{rev}.
Common challenges for cameras in achieving accurate human inactivity detection include dealing with complex background noise, coping with fluctuating lighting conditions (including low environmental lighting and abrupt changes, such as TV light flickering), meeting high sensitivity requirements for detecting small body movements (e.g., finger movements), and distinguishing human movements from non-human movements (e.g., pets).
{The method proposed below can cope with these difficult issues, as demonstrated below.}

\section{Unobtrusive Monitoring Methodology}\label{sec4}

{
The goal is to detect inactivity accurately and sensitively within a home environment where the resident spends a significant amount of time. This environment may include areas for reading, resting, or watching television. 
An inactivity event is defined as the absence of movement in any body part for a duration exceeding one second\footnote{{While ignoring tiny pauses (shorter than one second) between consecutive motions, as these might be inaccurate due to the limited frame rate (3 -- 5 fps).}}.}
{This allows us to record motionless periods relevant for data analysis, particularly in creating long-term mobility profiles.}
{Inactivity events initiate inactivity monitoring, which may reset upon the detection of motion, or may trigger an alert if inactivity persists for too long.}
Depth-based foreground extraction and color-based motion detection are used to detect if the person has stopped moving.

(i) Foreground detection\\
When monitoring people sitting in a room, the background can be complex, and the subjects can be in the region actively or inactively for long periods. A robust foreground detection method is applied. 
A background model is first constructed from a series of depth frames for the first few seconds when no one is present in the view. Each depth frame is smoothed by a median filter. Then, for each new depth frame, foreground pixels are detected by comparing them with the background model using a non-parametric method \cite{b30}. The non-parametric method fits a probability density function to the depth values at each pixel over a time window and detects changes. For every pixel at time t, the probability density function that this pixel has a depth value $d_t$ is calculated relative to previous background depth values $d_i$ at the same locations in  $n$ recent frames, as:
\begin{equation}
    Pr(d_t) = \frac{1}{n} \sum_{i=1}^{n}  {K_\sigma}(d_t - d_i),
    \label{eq6}
\end{equation}
where $K$ is a 1d Gaussian { with parameter $\sigma$}. The pixel is considered as a foreground pixel if the probability is less than the threshold. Each pixel is compared with several pixels at the same location in several recent background frames to enhance the robustness of foreground detection to small noise or vibrations in the background (e.g., leaf motion \cite{b30} or depth estimation errors). 
To speed up the calculation on a small computer {by avoiding the exponential computation for each pixel, 
{the log of} \eqref{eq6} is approximated by the quadratic function:
\begin{equation}
\begin{aligned}
f(d_t) = -\log \sqrt{2\pi\sigma^2} - \frac{1}{2n\sigma^2}\sum_{i=1}^{n}   {(d_t - d_i)^2}.
\end{aligned}
\end{equation}

The foreground (matrix) is derived as                          
\begin{equation}
    FG_t = {D}_t ^{ (f({d}_t) < \rho )}.
\end{equation}
In the implementation, $\rho$ is set to $-6.907$,} and $\sigma$ is set to $\frac{M}{0.68\sqrt{2}}$ \cite{b30}, where $M$ is the mean of the absolute value of pixel differences from successive depth frames in the background model. Once the foreground pixels are detected, the traditional post-processing procedures (size filter, tracking, and open processing) are applied to the foreground mask to remove noisy areas. The two most recent foreground regions are also saved (as a binary mask for privacy) for reference in the following misdetection suppression process.

Background frame pixels are selectively updated {for every loop}, excluding foreground pixels, since some observed body parts may not have any motion for a long time and should not be updated to the background for inactivity detection purposes. The latest background frame is updated and added to the background model sequence, and the oldest frame is removed. For the latest background frame at time $t$, the foreground area is not updated, it is directly copied from the background frame at time $t-1$,
whereas the previous background area is updated with the corresponding pixels from the current depth frame $D_t$, as:
\begin{equation}
   {BG}_t = {BG}_{t-1}^{\left( FG_t\right)} +  {BG}_{t-1}^{ \left( \neg FG_t \right) } \cdot  (1-\alpha)  +  {D_t}^{\left( \neg FG_t\right)}  \cdot  \alpha ,
\end{equation}
where $\alpha$ is the background update ratio.

\begin{figure}[h] 
    \centering 
    \includegraphics[width=.9\columnwidth]{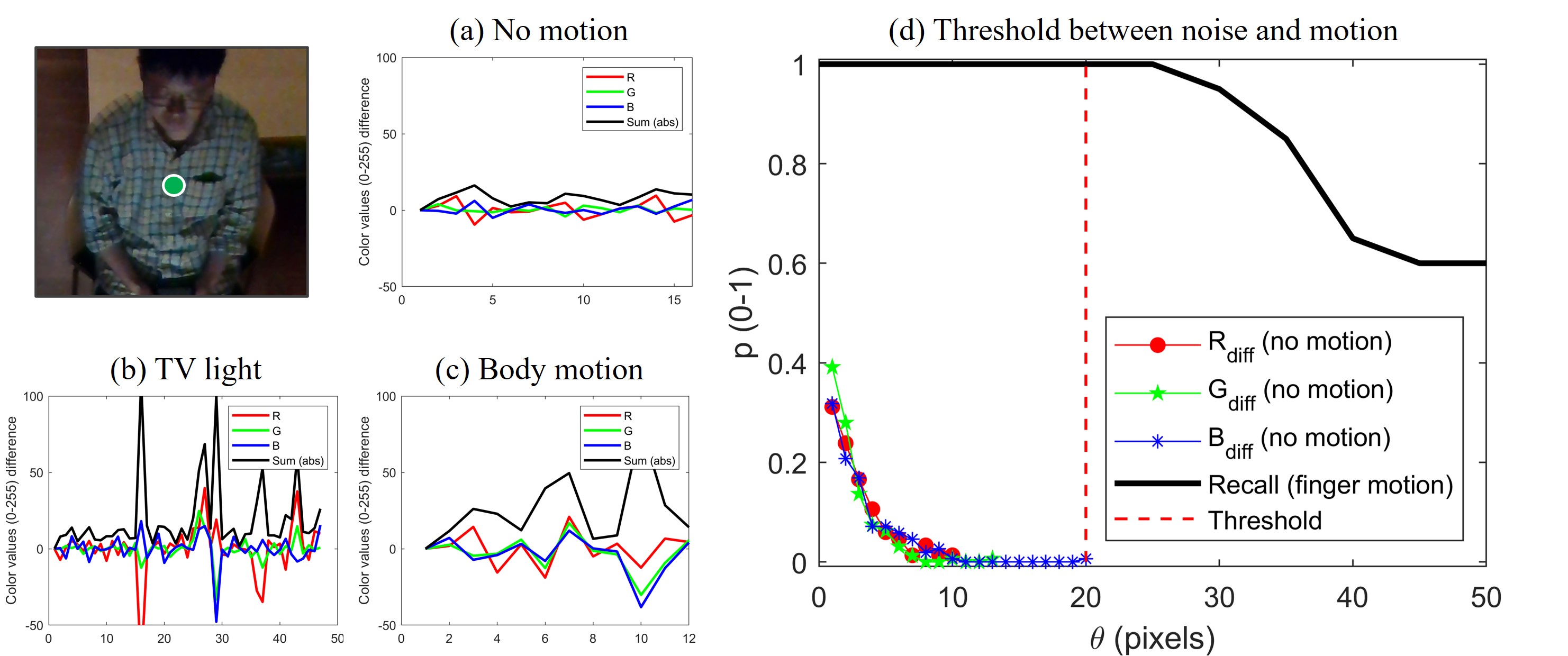}
    \caption{{RGB value difference of one color pixel (selected on the human chest) in consecutive frames under the following conditions: (a) no human motion under constant low light; (b) same environment, without human motion but with TV light changes; (c) human movement only. (d) The threshold for distinguishing noise (a) from true body motion (c), see main text for details}. (Scene Luma Rec. 601 \cite{rec601} $Y'$ is 20.5). \label{Fig.4}} 
    \end{figure}

(ii) Motion detection \\
Motion is detected using a color difference method, by subtracting the color of each pixel in the current frame from the color of the corresponding pixel in the previous frame, taking the absolute value, and comparing it to a threshold.
This basic method is sensitive to detect small true motions of the human body (e.g., fingers) and insensitive to slow changes in natural light; however, it is susceptible to small image artifacts and noise. For example, sudden lighting changes when watching TV at home, especially at night when the room light is dim. Sudden changes in TV light reflected on the human body can be mistaken for movement. 

{Fig. \ref{Fig.4} shows an example of color value changes of a pixel in a video that was recorded under the low light condition of a person in the living room watching TV:
(a) In the absence of human motion under constant lighting, pixels at the same image location have very small differences in color values between frames;
(b) In this sort of environment, when the TV light changes, peaks of the pixel color difference in one color channel often appear and disappear in a short period;  
(c) If it was true human motion, the pixel difference usually showed large peaks in all color channels and lasted longer.}

Therefore, to enhance robustness when classifying the changes as human motion, the change of intensities in all channels of the color pixels should be above a threshold, as:
\begin{equation} \label{eq5}
mv = FG \: \left[\mathbf{R}_{\rm{diff}}\geq\theta \:\cap\: \mathbf{B}_{\rm{diff}}\geq\theta \:\cap\: \mathbf{G}_{\rm{diff}}\geq\theta\right],
\end{equation}
where $mv$ is the movement pixels and $FG$ is the foreground region, $\mathbf{R}_{\rm{diff}}$ is the { absolute value of the} difference of pixel values between consecutive frames. 

The distribution of the pixel value differences in condition (a) in RGB channels over 300 frames was calculated. Since there is no movement, the difference value is considered as noises (ideally it should be 0). Moreover, in low lighting, the noise is usually larger than in good well-lit environments (i.e., upper bound of such noise). $\theta$ was increased to find the lower bound of losing any small movements (recall of 20 times finger movements, as illustrated in Fig. \ref{Fig.6}), and choose the proper threshold $\theta$ that can remove the noise while keeping the true small motions.

{
In the experiments,   a single threshold $\theta$ for all RGB channels is set at 20 (pixel value range 0-255). 
This threshold corresponds to the upper bound of image noise typically observed in low-light conditions.
Fig. \ref{Fig.4} (d) shows the distribution of pixel value differences of the RGB channels in the no-motion condition over 300 frames from both low-light and well-lit environments. Since there's no movement in these conditions, the differences ideally should be 0, with the values of this distribution indicating the noise level. 
Note that noise levels are higher in low-light conditions ($m$ = 2.23, $\sigma$ = 2.29) compared to well-lit environments ($m$ = 1.75, $\sigma$ = 1.27).
To identify the threshold where motion detection starts to miss small movements, 
The detection of small finger movements (over 20 trials) was investigated. 
From this, the detection threshold was increased to $\theta_{max}$
}
{$ = 25$, }
{to effectively remove noise while preserving true small motion detections. }

{A temporal median filter with a window size of 5 frames was then used to mitigate the effects of TV flickering and other sudden illumination changes. 
The temporal length of TV flickering was calculated across 40 trials. 
The average temporal length was 1.6 frames ($\sigma$ = 0.74). This suggests that sudden illumination changes are usually less than 0.5 seconds.}
{Furthermore, considering the spatial and temporal size of the true motion of the human body (i.e., larger than a square centimeter, usually more than 0.5 seconds), a 2d spatial median filter (5 $\times$ 5 pixels) is applied to the foreground regions to further eliminate tiny isolated noise pixels.}

\begin{figure}[h]
    \centering
    \subfigure[Before region growing]{
        \includegraphics[width=.3\columnwidth]{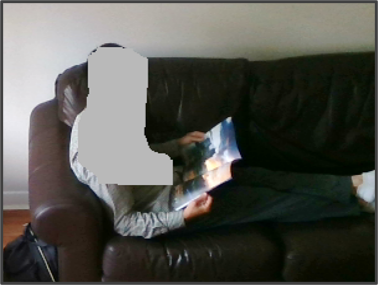}
    }\  \   \  \   \  \   \  \   \  \  \  \  \  \  \  \  \  
    \subfigure[After region growing]{
	\includegraphics[width=.3\columnwidth]{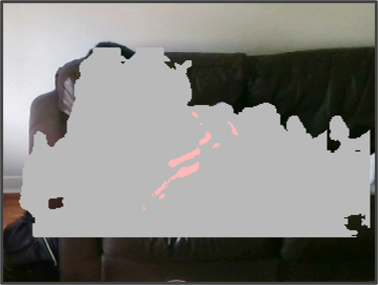}
    }
    \caption{A person sitting on a couch. The grey area shows the foreground detected using the depth map (a) before the region growing and (b) after the region growing. Real human motion (red) is not detected in the incomplete foreground before the region growing.
}
    \label{fig.5}
\end{figure}

(iii) Misdetection suppression \\
The motion regions detected using the depth-based method in (i) above are often disconnected when the person adopts some poses, such as reclining or lying on a couch.
This is because, in these poses, the depth values of the background (e.g., couch) are very close to the depth values of the body parts, and most detection methods that calculate the difference between foreground and background depths may not be able to distinguish the small differences given inaccurate camera measurements ($<2\%$ depth error at 2 meters). An example is shown in Fig. \ref{fig.5}.

To deal with this incomplete foreground, the detected foreground is grown by referencing its depth and spatial locations on the image.
An enlarged bounding box (10\% of image width) is set around the foreground area to grow within.
The mean ($m$) and standard deviation ($\sigma$) of foreground depth values are calculated as growth reference values.
For any pixel 
{adjacent to a foreground region}
in the enlarged bounding box, if its depth value $d$ is close to the foreground mean $m$, as:
\begin{equation}
   \lvert d - m\rvert < 2.8 \sigma,
\end{equation}
then this pixel is treated as a foreground inlier and will grow into the foreground.
If the nearby background regions are not close to the foreground, there will be no region growth.
If multiple foreground regions are detected, only the regions that are close to the most recent foregrounds in size and depth are grown.

\begin{figure}[t]
    \centering
    
    \subfigure[Pet motion only]{
        \includegraphics[width=.23\columnwidth]{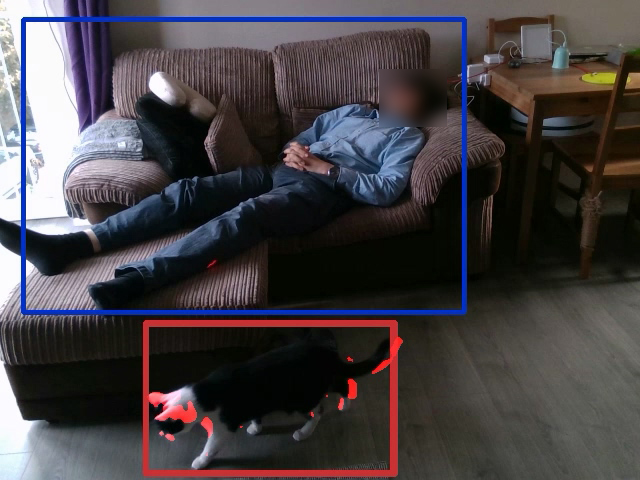}
    }\  \   
    \subfigure[Removed pet motion]{
	  \includegraphics[width=.23\columnwidth]{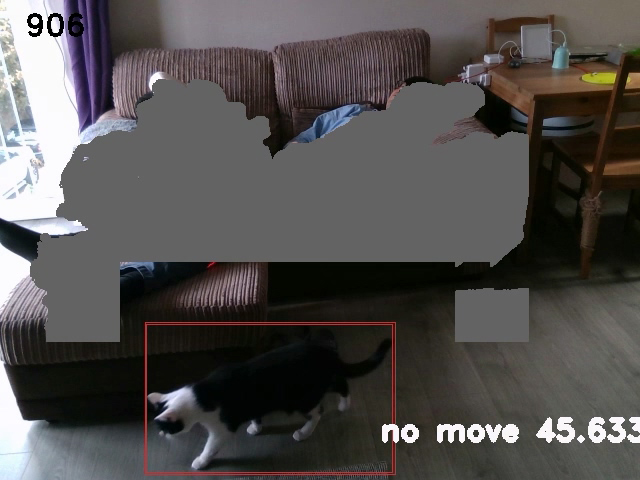}
    }
    \subfigure[Failed to detect pet]{
        \includegraphics[width=.23\columnwidth]{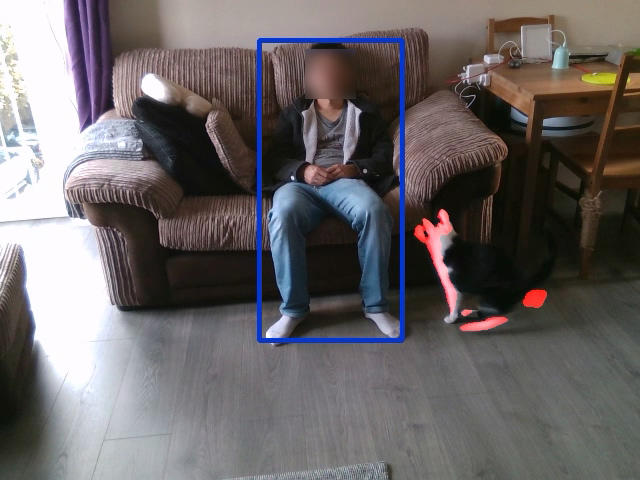}
    }\  \   
    \subfigure[Failed to remove pet motion]{
	\includegraphics[width=.23\columnwidth]{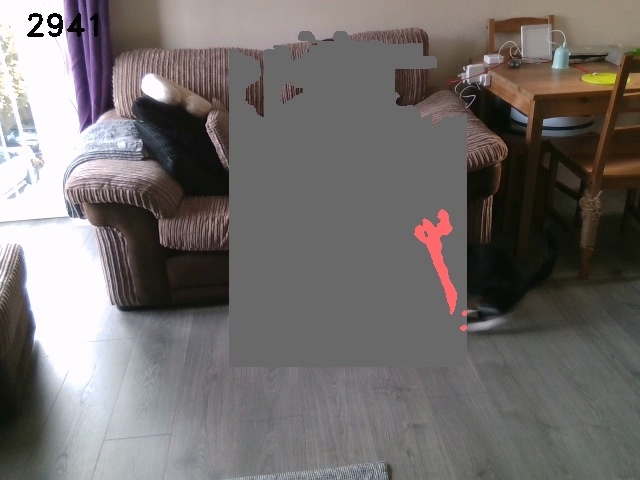}
    }
    \caption{{Object detector helps remove pet motion. (a) Pet movement occurs while humans are inactive. (b) The pet is successfully detected, and the relative region is removed from the foreground, allowing the human inactivity count to continue. (c) The pet is not detected by the object detector. (d) Pet motion is not removed, resulting in the cessation of the human inactivity count.}}
    \label{fig.c}
\end{figure}

{Human/Pet detector}: 
Large objects in the background, such as chairs or tables that may shift when a person leaves their seat, are sometimes detected during monitoring and remain in the foreground. These objects cannot be eliminated through selective background updates.
To address this, an object detector (pre-trained YOLOv5, including humans and pets) is employed. 
Since object detectors are computationally expensive for real-time processing and may not always perform reliably with complex backgrounds and with various human poses, the detection occurs every 10 seconds.
The system aggregates multiple observations and votes for human or non-human. 
After a minute, if the vote result ($\geq80\%$) indicates `not human', the system updates the foreground region to become part of the background. 

{Pets, which are often present in the homes of single older adults, may move around humans even when the human is motionless, and this can introduce unwanted motion detections.
When both humans and pets are present, the detector runs frame by frame, subtracting the detected pet region (bounding box) from the foreground region. Motion is then detected only within the remaining foreground region, effectively excluding pet motion, as shown in Fig. \ref{fig.c}.}

\subsection{Behavior Statistics and Models} \label{subsec3}

{For inactivity detection, if no human motion is detected in the foreground, a timer will start counting the inactivity period in seconds.
Once motion is detected in the foreground (and excluding the pet region), or if the foreground is recognized as non-human, the timer will reset.}
Periods of inactivity ($\geq1$ second) and their occurrence times are saved in a log. Median, maximum, and minimum inactivity periods, by time of the day, are extracted from the log data.

Fitting distributions to the data was investigated to enable long-term comparison within the same subjects or among multiple subjects. The number of movement occurrences over a period of time (e.g., 1 minute, 1 hour) can be modeled with a Poisson distribution if one assumes that movements occur independently. Then the inactivity period between any consecutive movement events can be modeled with an exponential distribution as:
\begin{equation}
f(x; \lambda) = \lambda e^{(-\lambda x)}   ; x>=0 ,
\end{equation}
where the maximum likelihood estimate of the parameter $\lambda$ is the inverse of the mean of the data as $\lambda_{mle} = 1/ \bar{x}$.

\section{Results}\label{sec5}
\subsection{System Configuration} \label{subsec4}

{
The RGB-D camera is an Intel RealSense D415 \cite{b35} with an ideal range of 0.5 meters to 3 meters. A USB cable connects the camera to the processor, a Jetson Nano \cite{b36}, which requires a minimum of 4.75 volts and operates on as little as 5 watts.}
The overall dimensions of the system are less than 20 cm $\times$ 10 cm $\times$ 15 cm (excluding power cables).
The camera captures color and depth imagery at a resolution of 640 $\times$ 480. The images are processed by the processor in real time, extracting the foreground and motion features.
For the detection task, both depth and color images are used. 5 fps was observed when no one was present, and 3.5 fps was observed when someone was in view. 
{The process is efficient and has no lag; inactivity events will be reported within 1 second (see details in Section \ref{subsec5}: Temporal Sensitivity). }

{\textbf{Privacy-preserving}: The monitoring is camera-based, but the video frames are only temporally loaded in the RAM for on-the-fly analysis and then discarded. No images or videos are stored or transferred anywhere.} 
Currently, the device has no internet connection, so there is no risk of hacking or people viewing the subject, not even the researchers {(see the detailed discussion in Section \ref{secextra}).} 
All that is stored about the recorded events are when someone is seen at the monitored location, and how often they move. 
Detected events are stored anonymously in the form of text event logs.
{It is assumed that these logs will be analyzed on the home device, with the results available to visiting health workers.}

\subsection{Experimental Methodology}
{As an anonymous and privacy-preserving monitoring method, obtaining the ground truth for evaluation is challenging. Therefore, controlled experiments were conducted in a variety of environments to evaluate the accuracy of the system. Images were saved, and the ground truth was manually labeled. 116 videos were recorded across 12 indoor scenarios. This accuracy evaluation of the system included the following aspects: (\Romannum{1}) Motion detection, 
(\Romannum{2}) Spatial and temporal sensitivity, (\Romannum{3}) Robustness in low light conditions and against TV light, and (\Romannum{4}) Robustness against pet motion.
}
{Two state-of-the-art motion detection methods are compared. One is a kinetic-based pose estimation method, ViTPose \cite{vit}; the other is an optical flow-based method, RAFT \cite{raft}. Both methods are pre-trained on transformer-based deep neural networks. 
The pre-trained coefficients of the networks were used and then tuned both methods to small motion sensitivity levels while keeping the upper bound of the noise threshold to remove as much noise as possible. The results are presented in Section \ref{subsec5}.}

{Subsequently, our system was tested in a real-life deployment to capture the inactivity patterns of older adults. The monitoring system was deployed in four older adult households within the community, as an ethically approved study. In this home monitoring, no video or image data was saved, and no ground truth of the detected inactivity events was obtained. The results are presented in Section \ref{subsec6}.}

\subsection{Accuracy Evaluation}\label{subsec5}

\begin{figure*}[t]
    \centering
    \includegraphics[width=.95\textwidth]{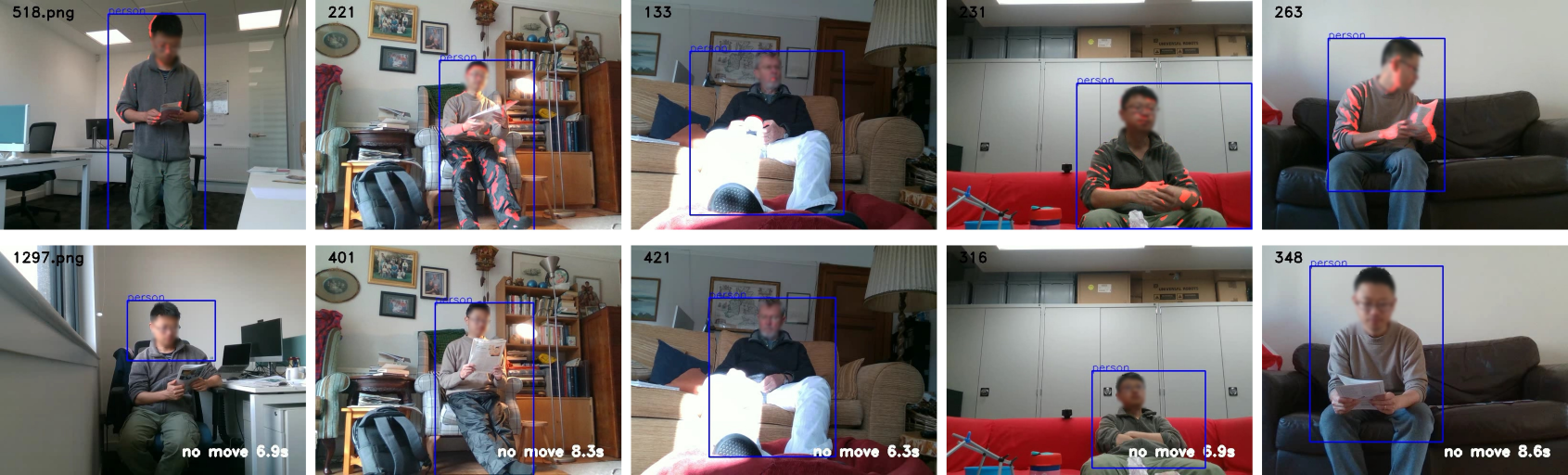}
    \caption{Examples of inactivity detection evaluation in different scenarios (front view in office, lab, and homes). The top row shows people in motion, and the bottom row shows people remaining inactive. Detected humans are marked with blue boxes, and areas with detected motion are highlighted in red pixels.}
\label{fig.15}   
\end{figure*}

\textbf{(\Romannum{1}) Motion detection}.
To evaluate the detection accuracy, video capture sessions with controlled motion and no-motion behaviors were conducted.
A person started with continuous movement (about 30 seconds), then went to no movement at all (about 10 seconds), and then started continuous movement again (about 20 seconds).
This process was repeated 70 times and 70 short videos were recorded in laboratories, offices, and homes, as illustrated in Fig. \ref{fig.15}.
The ground truth for these events, i.e., humans appear/disappear, movement starts/ends) is labeled manually as frame numbers in the videos.

Four types of errors are evaluated frame-wise.
Human detection false positives (HuFP) and false negatives (HuFN), and motion detection false positives (MoFP) and false negatives (MoFN), are shown in Table \ref{tab3}.
\footnote{From seventy video recordings with controlled motion/no-motion behaviors.}
To calculate the motion FN, it was assumed that during continuous movement periods, all frames are motion positive, although there are sometimes short pauses ($\ll$1s) in transitions of body movement. During periods of complete motionlessness, all frames are labeled as motion-negative.

The results show that MoFP has a low error rate in all scenarios - from $0.48\%$ to $0.00\%$ when the temporal matching tolerance increases from $\pm$0 to $\pm$5 frames for the beginning/end moment of the completely no motion periods.
This shows that, although a few frames of mismatch are tolerated at the beginning/end of the inactivity period, every no-motion frame is correctly detected in periods of inactivity, where the detection is insensitive to noise, i.e., the detection method can filter out all types of background noise or fake motion caused by lighting changes.
MoFN is at 4.31\% to 3.66\% ($\pm$0 to $\pm$5), indicating that several frames of true motion during continuous motion periods were missed.
HuFP is 1.46\% to 0.97\% ($\pm$0 to $\pm$5) in all frames and HuFN is 3.59\% to 3.23\% ($\pm$0 to $\pm$5), which shows that the human detector (pre-trained YOLOv5) is more likely to miss real people than detecting background regions as humans in these experimental environments.
{In this experiment, there are many short (mostly $<$ 1 second) periods of no detected motion during the intervals labeled in the ground-truth as having motion. These are the frame-wise MoFN errors. However, when considering the motion instance level that fuses motion across these brief gaps, all motion and non-motion instances are correctly detected.}

\begin{table}
    \centering
    \caption{Error rates for inactivity detection.  The error types are: 
    \textbf{HuFP}: detected background region as human, which usually starts the inactivity time counter, resulting in false inactivity events.
    \textbf{HuFN}: missed real human detection, which causes the person to be updated into the background and reset the timer, resulting in missing inactivity events;
    \textbf{MoFP}: detected no motion as true motion, which falsely resets the timer, causing both missed and fewer true long-period inactivity events; 
    \textbf{MoFN}: missed true human motion which causes a failure to reset the inactivity time counter and extends the count period, causing false long-period inactivity events.
    {Because of potential mislabeling in the ground truth, a given number of frames of mismatch Tolerance is allowed in the bottom three rows.}
    }
    \label{tab3}
    \begin{tabular}{ccccc}
    \toprule
    Tolerance  & HuFP & HuFN & MoFP & MoFN \\
    (frames) & ($10e{-2}$) & ($10e{-2}$) & ($10e{-3}$) & ($10e{-2}$)\\
    \midrule
    $\pm$0  & 1.46 & 3.59  & 4.8  & 4.31                   \\
    $\pm$1  & 1.31 & 3.47  & 1.7  & 4.10                 \\
    $\pm$3  & 1.15 & 3.31  & 0   & 3.82                   \\
    $\pm$5  & 0.97 & 3.23  & 0  & 3.66            \\
    \bottomrule
    \end{tabular}
    \end{table}

\begin{figure}[t] 
    \centering 
    \includegraphics[width=.7\columnwidth]{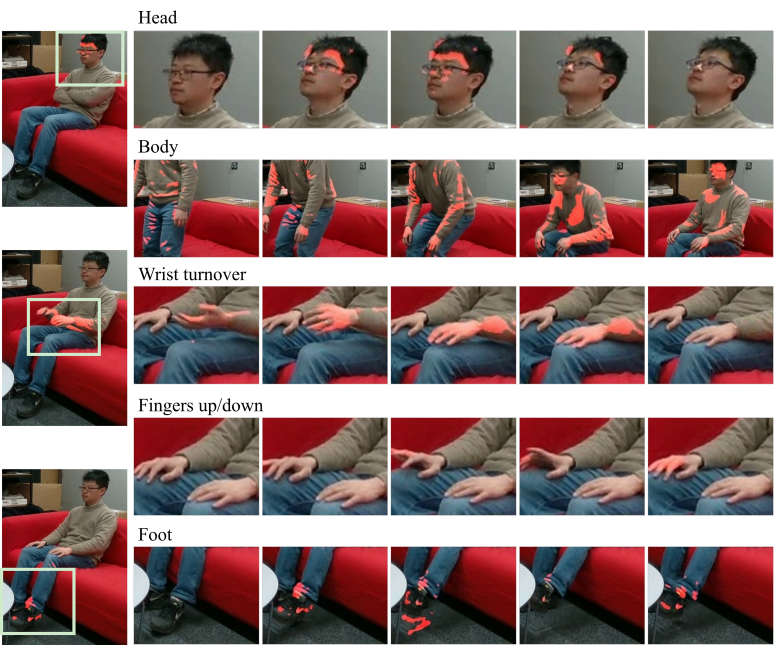} 
    \caption{Examples of spatial resolution of human motion detection under good lighting conditions in the laboratory (side view, Luma $\:Y'\:107$). Five body movements were performed, 20 repetitions each (the head 40 repetitions for 20° rotation). Recall rates: Head (1.0), Body (1.0), Wrist (1.0), Fingers (1.0), Feet (1.0).} 
    \label{Fig.6} 
\end{figure}

\begin{table}[h]
    \centering
    \caption{{Recall (R) and Precision (P) Under Different Body Motion Patterns (Motion Detection Accuracy)}}
    \begin{tabular}{lcccccc}
      \toprule
            \multirow{2}{*}{Lab ($Y'$107)} & \multicolumn{2}{c}{ViTPose\cite{vit}} & \multicolumn{2}{c}{RAFT\cite{raft}} & \multicolumn{2}{c}{MISO} \\ 
            & R$\uparrow$ & P$\uparrow$ & R$\uparrow$ & P$\uparrow$ & R$\uparrow$ & P$\uparrow$ \\ \midrule
        body & \textbf{20/20} & \textbf{20/20} & \textbf{20/20} & \textbf{20/20} & \textbf{20/20} & \textbf{20/20} \\ 
        head & 25/40 & 25/25 & \textbf{40/40} & \textbf{40/40} & {38/40} & {38/38} \\ 
        wrist & \textbf{20/20} & \textbf{20/20} & \textbf{20/20} & \textbf{20/20} & \textbf{20/20} & \textbf{20/20} \\ 
        finger & 0/20 & 0/0 & {19/20} & {19/19} & \textbf{20/20} & \textbf{20/20} \\ 
        foot & \textbf{20/20} & \textbf{20/20} & \textbf{20/20} & \textbf{20/20} & \textbf{20/20} & \textbf{20/20} \\   \bottomrule
    \end{tabular}
    \label{tabpattern}
\end{table}

\textbf {(\Romannum{2}) Spatial and temporal sensitivity.}
{
To evaluate the spatial sensitivity of our method, a subject performed five physical movements repeatedly and was monitored by a side-view camera 2.6 meters away. Fig. \ref{Fig.6} illustrates the sensitivity of detecting human movements under different motion patterns. }
{The movement ground truth was}: body, wrist, finger, foot, each repeated 20 times (10 to the left, 10 to the right), and 40 head movements (20° each in four directions).
The results in Table \ref{tabpattern} demonstrate that our method has good sensitivity, with a true positive rate of 1.0 for detecting four motion types, including both large movements (such as sitting) and small movements (such as finger lifting). 
The true positive rate for detecting small head motion is 0.95, where all head motions were detected, however, few (2 out of 40) happened too quickly (2 frames or less) and were removed by the temporal filter.
{Whereas the ViTPose pose-based model \cite{vit} missed all of the finger motions and 25\% of the head motions since it only estimated the main body joints, and was not as fine-grained as the RAFT optical-flow-based method \cite{raft} nor our proposed method.}

To evaluate the temporal sensitivity, 30 video clips of inactivity events ($\geq5s$) were collected in a well-lit laboratory, manually noted the beginning and end moment of each event as ground truth, and then compared these with the detected log. The temporal accuracy of the inactivity detection result is $\pm$1.4 frames (at 3 to 5 fps), {which indicates that the inactivity events will be reported promptly within 1 second.}

\begin{figure*}[t] 
\centering 
\includegraphics[width=\textwidth]{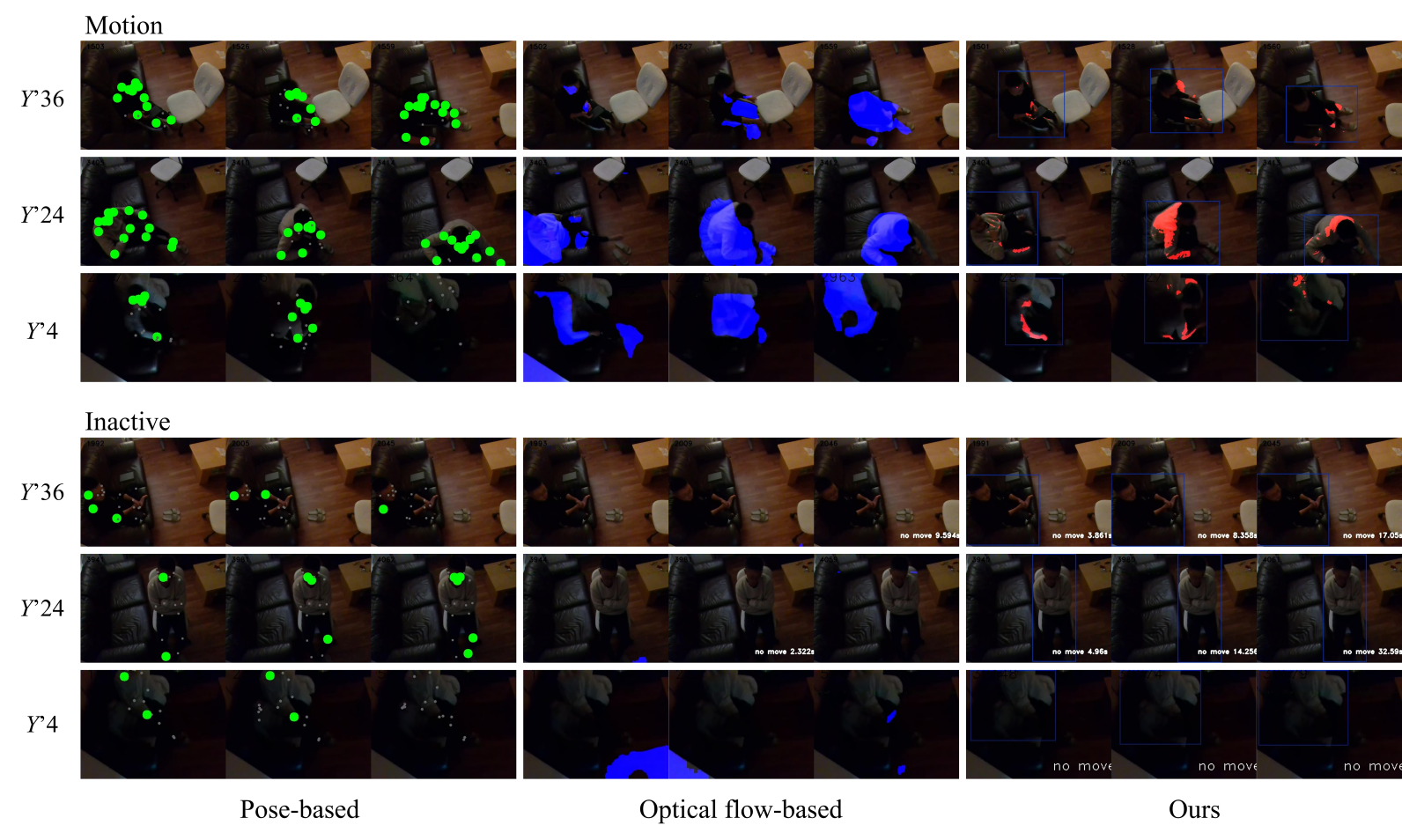} 
\caption{{Performance comparison in low lighting conditions (top view). From left to right: VitPose\cite{vit}, RAFT\cite{raft}, and our method. Detected motion is shown in green, blue, and red, respectively, for the methods.}} 
\label{Figlow} 
\end{figure*}

{\bf (\Romannum{3}) Robustness in low light conditions and against TV light flicker}. 
Motion detection assessments were then performed under varying lighting conditions.
The cameras recorded the subjects' activities with an oblique view at 1.5 meters, as illustrated in Fig. \ref{Figlow}.
The detection performance is then evaluated by the number of motions missed given 20 ground-truth movements under each lighting condition.
The lighting condition has 5 levels from high to low, as daylight (Rec. 601 Luma $Y'$ 97), night light ($Y'$ 75), low light ($Y'$ 36), dim light ($Y'$ 24), and dark ($Y'$ 4). 
The results (Table \ref{tablow}) show that under the first four lighting conditions ($Y'$ 24 to 97), 1 or 2 out of the 20 movements were not detected (error rate 5\% to 10\%); whereas in the darkest environment ($Y'4$), the misdetection rate of true motions was 40\% (8 out of 20 events were not detected). This indicates the limitation of our detection method under extremely low-lighting environments.

\begin{table}
    \centering
    \caption{{Recall (R) and Precision (P) Under Different Lighting Conditions (Motion Detection Accuracy)}}
    \begin{tabular}{lcccccc}
    \toprule
        \multirow{2}{*}{Room Lighting}  & \multicolumn{2}{c}{ViTPose\cite{vit}} & \multicolumn{2}{c}{RAFT\cite{raft}} & \multicolumn{2}{c}{MISO} \\ 
        & R$\uparrow$ & P$\uparrow$ & R$\uparrow$ & P$\uparrow$ & R$\uparrow$ & P$\uparrow$ \\ \midrule 
        day ($Y$'97) & 16/20 & 16/17 & 16/20 & 16/17 & \textbf{19/20} & \textbf{19/19} \\
        night ($Y$'75)  & \textbf{19/20} & \textbf{19/21} & 17/20 & 17/17 & 18/20 & 18/18 \\
        low ($Y$'36) & 20/20 & 20/200* & 19/20 & 19/39 & \textbf{19/20} & \textbf{19/19} \\ 
        dim ($Y$'24) & \multicolumn{2}{c}{nan\^} & 15/20 & 15/210* & \textbf{18/20} & \textbf{18/18} \\
        dark ($Y$'4) & \multicolumn{2}{c}{nan\^} & \multicolumn{2}{c}{nan\^} & \textbf{12/20} & \textbf{12/12} \\ \bottomrule
        \multicolumn{7}{l}{\^{} : \footnotesize Excessive noise led to many false positives.} \\
        \multicolumn{7}{l}{* : \footnotesize Approximated by the average ratio of the number of FP instances to each TP instance.}\\
    \end{tabular}
       \label{tablow}
\end{table}

    \begin{table}
        \centering
        \caption{{Human Motion False Positive(FP) Under TV Light Flickering Conditions}}
        \begin{tabular}{lcccc}
        \toprule
            \multirow{2}{*}{Lighting}   & \multirow{2}{*}{TV Dis.}  & ViTPose\cite{vit}  & RAFT\cite{raft} & MISO \\
            &  & FP$\downarrow$ & FP$\downarrow$ & FP$\downarrow$ \\  \midrule
             $Y$'90 & 0.7m & \textbf{0/20} & 3/20 & \textbf{0/20} \\ 
            ~ & 1.5m & \textbf{0/20} & 5/20 & \textbf{0/20} \\ 
             $Y$'41 & 0.7m & 3/20 & 6/20 & \textbf{0/20} \\ 
            ~ & 1.5m & \textbf{0/20} & 8/20 & \textbf{0/20} \\
             $Y$'14 & 0.7m & 14/20 & nan\^ & \textbf{0/20} \\ 
            ~ & 1.5m & 19/20 & nan\^ & \textbf{0/20} \\ 
             $Y$'4 & 0.7m & 15/20 & nan\^ & \textbf{0/20} \\ 
            ~ & 1.5m & nan\^ & nan\^ &  nan* \\ \bottomrule
            \multicolumn{5}{l}{\^{} : \footnotesize Excessive noise led to many false positives.} \\
            \multicolumn{5}{l}{* : \footnotesize The human detector failed in the dark environment.} \\
        \end{tabular}
           \label{tabtv}
    \end{table}

\begin{figure}[h] 
    \centering 
    \includegraphics[width=.7\columnwidth]{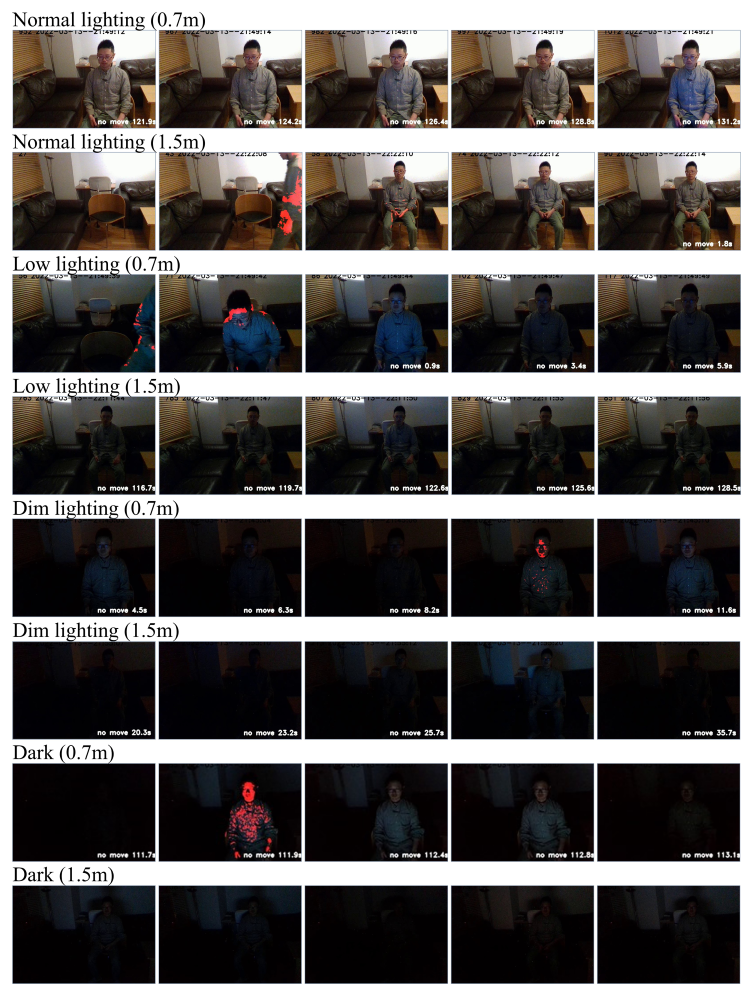} 
    \caption{Examples showing motion detection is robust to TV light changes at different person-to-TV distances (0.7m or 1.5m) and four different lighting conditions (top-to-bottom, Luma $Y'$ are 90, 90, 41, 41, 14, 14, 4, 4). False positive rate (top-to-bottom): 0/20, 0/20, 0/20, 0/20, 0/20, 0/20, 0/20, nan.
    } 
    \label{Fig.8} 
    \end{figure}

As mentioned previously, watching TV is a common activity for older adults, especially in dimly lit rooms, where the light from the TV can easily scatter colors onto people to affect the chromatic-based detection methods. To test the robustness of the motion detection against changes in TV lighting, an experiment was conducted under different room lights and at different distances from the TV. 
Examples are shown in Fig. \ref{Fig.8}.
Subjects sat at 0.7 meters and 1.5 meters from the TV under 4 different room lighting conditions and then changed the TV light by changing the TV channel, 20 times at each environment setting as ground. 
The results in Table \ref{tabtv} show that the false positive rates \footnote{Detection accuracy is represented by the false positive rate. False positives occur when the TV light reflected onto the body is classified as motion.} are 0\% in all settings, except in the darkest and furthest condition (1.5 m, Luma $Y'4$), where the person detector often failed to identify a person sitting in the scene.
Although a few changes in TV global illumination were detected (see Fig. \ref{Fig.8}), they were not classified as true motion by our method, as these changes only occurred briefly (less than 0.5 seconds and thus were removed by the temporal filter).
{On the other hand, the pre-trained deep neural network models both performed worse in low-light and TV flickering conditions compared to our method (see Table \ref{tablow} and Table \ref{tabtv}). The optical flow-based method exhibited significant false positives in non-human regions under low-light conditions, whereas the pose-based method could detect the human joints but also detected substantial jitters, even when the person remained inactive, as illustrated in Fig. \ref{Figlow}.}

{{\bf (\Romannum{4}) Robustness against pet motion}.
To test our detection method against pet motion, two subjects participated in three inactivity trials in a home scenario with a cat present (see Fig. \ref{fig.c}). Each inactivity trial lasted approximately 10 minutes, resulting in a total inactivity duration of about 30 minutes (10,898 frames). The object detector runs frame by frame to detect the cat's region and remove it from the foreground. }

The recall rate\footnote{The rate is calculated as the ratio of recalled inactivity seconds to the total inactivity seconds.} for the true human inactivity duration, is presented in Table \ref{tabcat}.
{Without removing the pet region, an average of 63.7\% of the true human inactivity duration was recalled. When the pet moved around, it introduced random motions that significantly affected the accuracy of human inactivity detection. With the removal of the pet region during inactivity detection, the recall rate for the true human inactivity duration improved to 82.4\% for our method.}
{For comparison, the flow-based method was easily affected by both pet and background movement, whereas pose-based methods detected only human joints and achieved comparable performance with our method.} The primary reason for this limitation was the pre-trained object detector's failure to detect the pet under certain conditions, such as specific poses or when the pet was moving quickly (resulting in motion blur), as illustrated in Fig. \ref{fig.c} (c) and (d). 
Enhancing the pet detector could help address this problem, but this falls outside the scope of the current study.


\begin{table}[h]
    \centering
    \caption{{Average Recall (\%) of Human Inactivity when Pets are Around}}
    \label{tabcat}
    \begin{tabular}{ccccc}
         \toprule
    \multirow{2}{*}{Trial} & {MISO (without} & MISO (with  & \multirow{2}{*}{ViTPose\cite{vit}}  & \multirow{2}{*}{RAFT\cite{raft}} \\
     & {pet removal)} & {pet removal)}  &   &  \\
       \midrule 
    T1 & 59.8\%                       & 74.6\%        & \textbf{94.30\%}    &         52.60\%            \\
    T2 & 69.4\%                      & 85.5\%         & \textbf{97.70\%}    &     47.90\%                 \\
    T3 & 60.3\%                       & \textbf{85.8\%}        &60.5\%     &       68.10\%                 \\
    \bottomrule
    \end{tabular}
\end{table}

\subsection{Field Study}\label{subsec6}

{The monitoring system was deployed in four households in the older adult community. Residents ranged in age from 65 to 80 years old, {with two male older adults and two female older adults.}
For each participant's house, the system captured data over a period of approximately three days. 
For the inactivity detection, all cameras were in the living room on a tripod, two observed a couch, one observed a chair, and the last one viewed both a chair and a couch.}
{In total, 23,393 log events of inactivity ($\geq$1s) were recorded.}

\begin{figure}[t]
    \centering
    \subfigure[]{
        \includegraphics[width=.45\columnwidth]{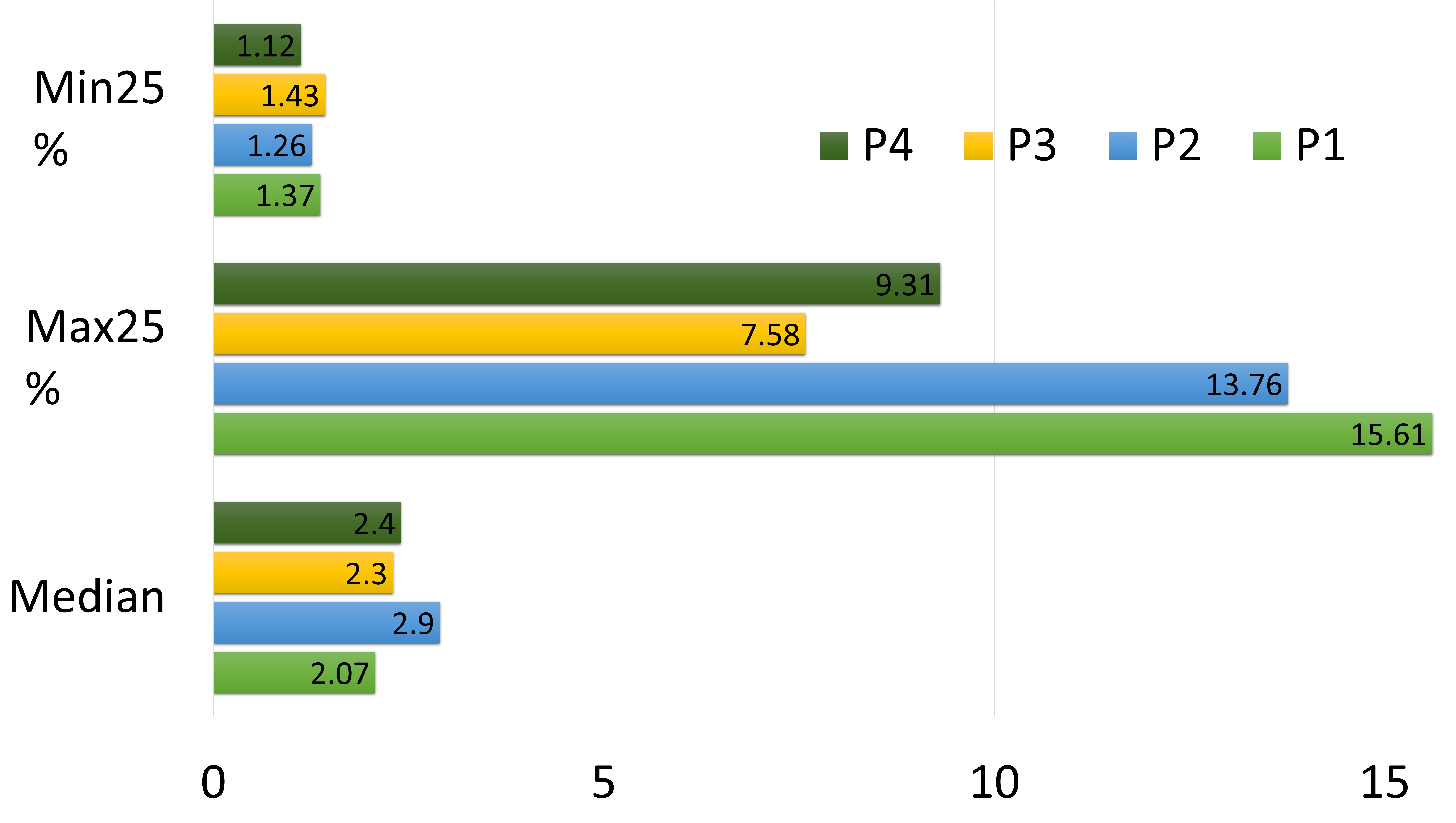}
    }\  \  \  \  \  \  \  \  \  \  \  \  
    \subfigure[]{
	\includegraphics[width=.45\columnwidth]{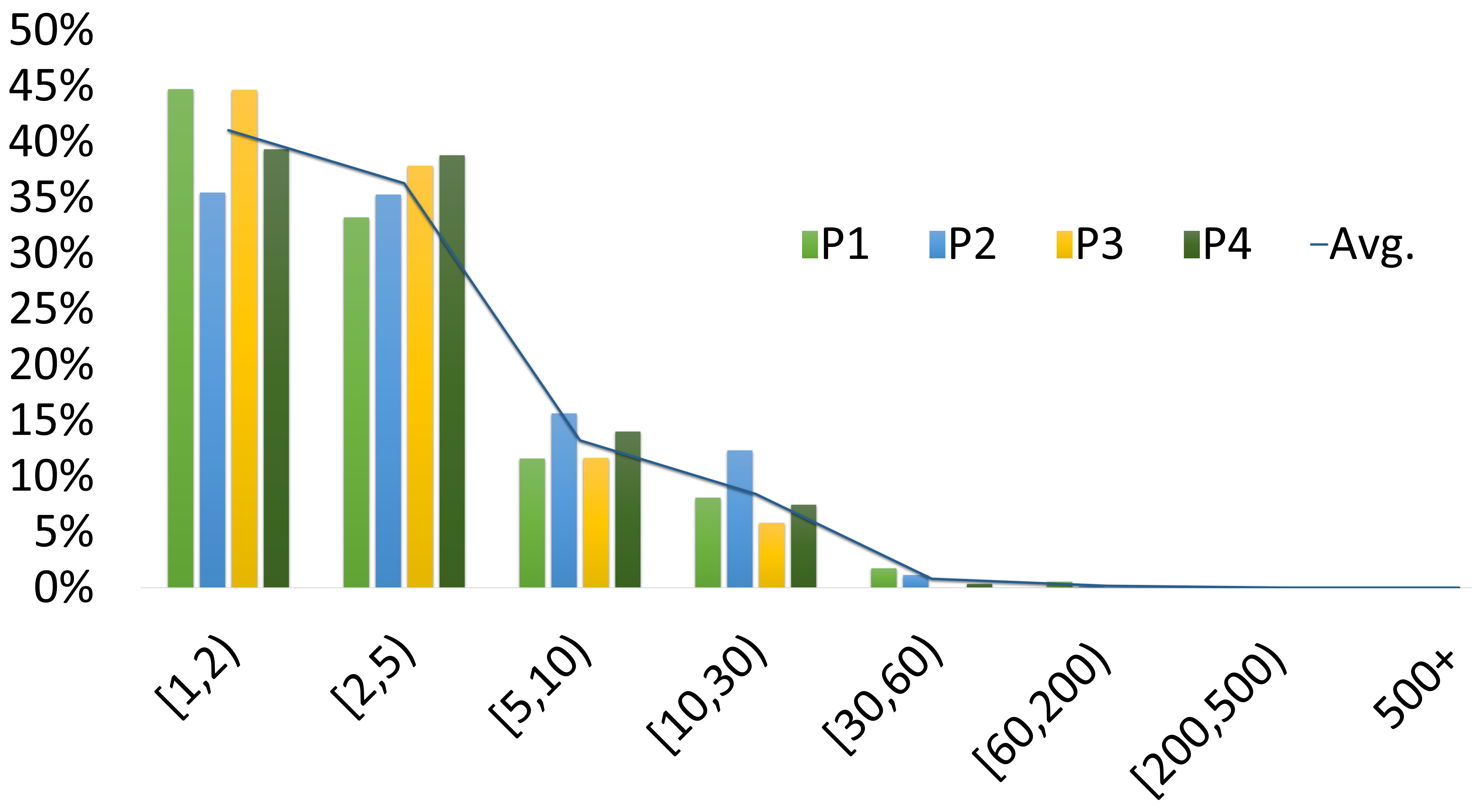}
    }
    \caption{Inactivity detection statistics for four participants. (a) Average inactivity time of each participant (Median, Max 25\%, Min 25\%). (b) Percentage of inactivity duration for each different time range (seconds).
}
    \label{fig.10}
\end{figure}

Table \ref{tab6} summarizes the inactivity statistics of four older adult participants on a chair or couch in the living room.
A median of 2.0 to 2.9 seconds for periods of non-movement was detected across all participants, with the minimum 25\% intervals of 1.0 to 1.5 seconds.
When the maximum 25\% periods of inactivity periods are considered, the duration varied between participants, ranging from 7.5 seconds to 15.6 seconds.
This demonstrated that a person tends to remain completely still for only very short periods of time when they are awake. Fig. \ref{fig.10} shows the statistics.

\begin{figure}[t]
    \centering
    \subfigure[Inactivity time of day]{
        \includegraphics[width=.43\columnwidth]{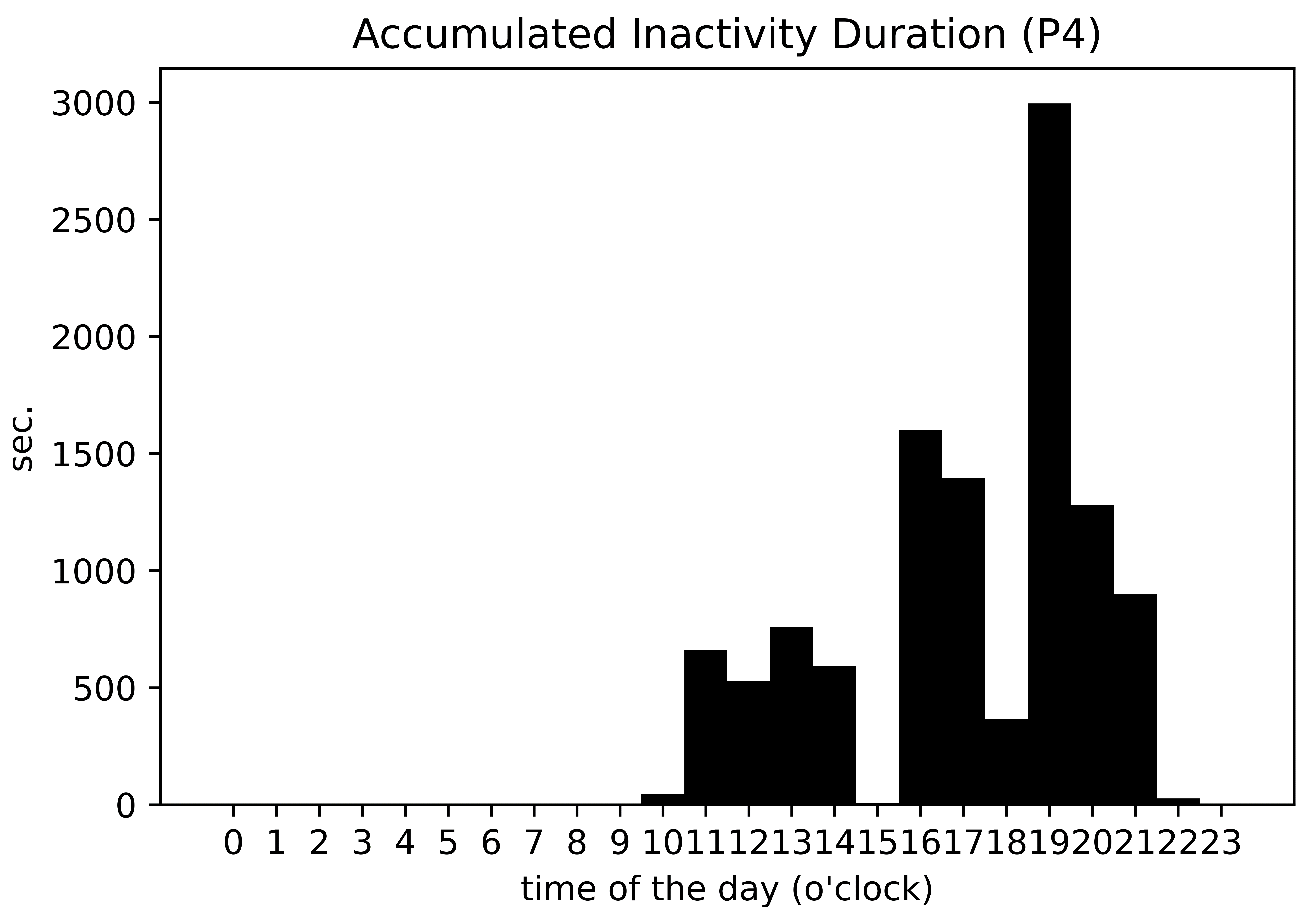}
    }\  \  \  \  \  \  \  \  \  \  \  \  
    \subfigure[Inactivity duration distribution]{
	\includegraphics[width=.43\columnwidth]{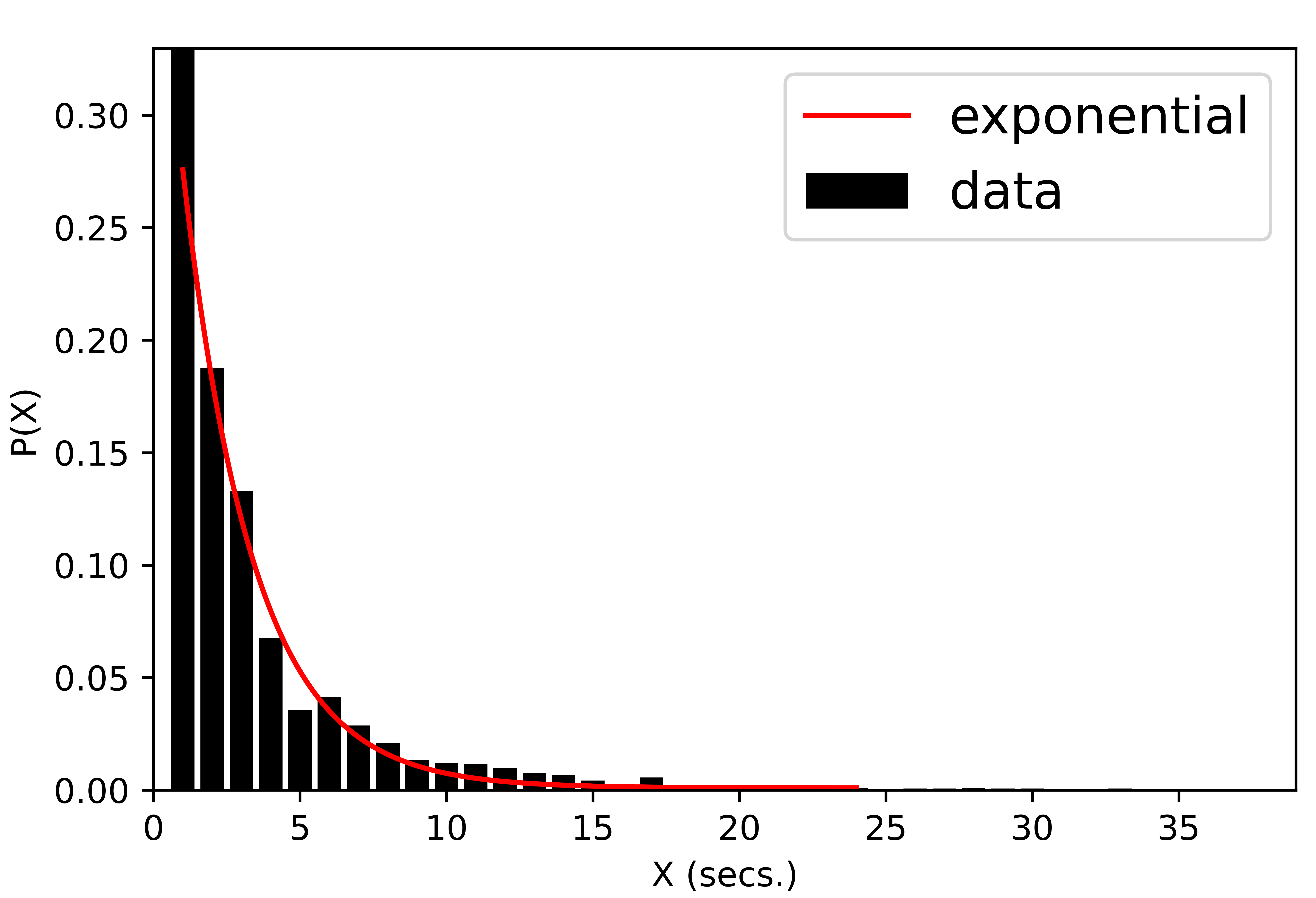}
    }
    \caption{Example of inactivity statistics (Older Adult Participant 4).
}
    \label{fig.11}
\end{figure}

\begin{figure}[t]
            \subfigure[Inactivity time-of-day]{
                \includegraphics[width=.27\columnwidth]{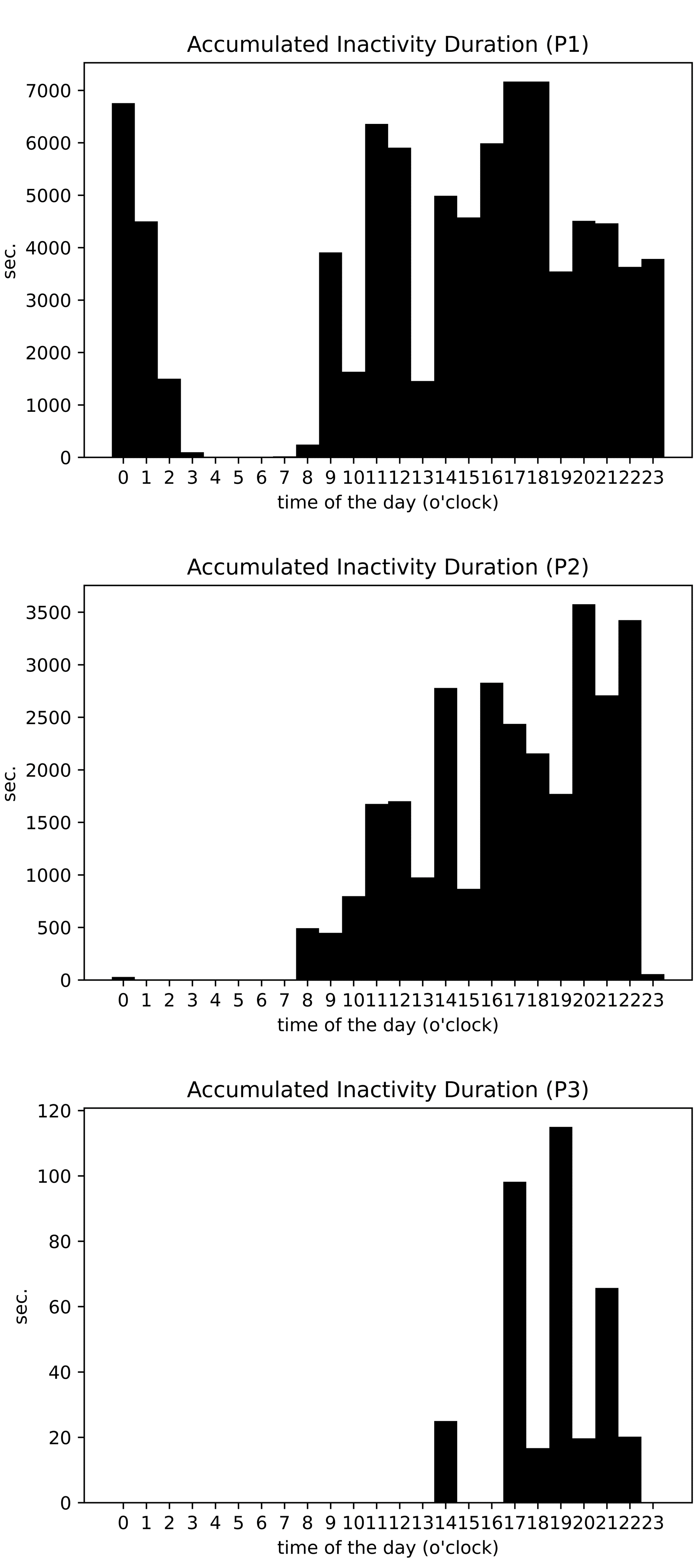}
            }\  \  \  \  \  \  \  \  \  \  \  \  \  \ 
            \subfigure[Inactivity distribution]{
            \includegraphics[width=.27\columnwidth]{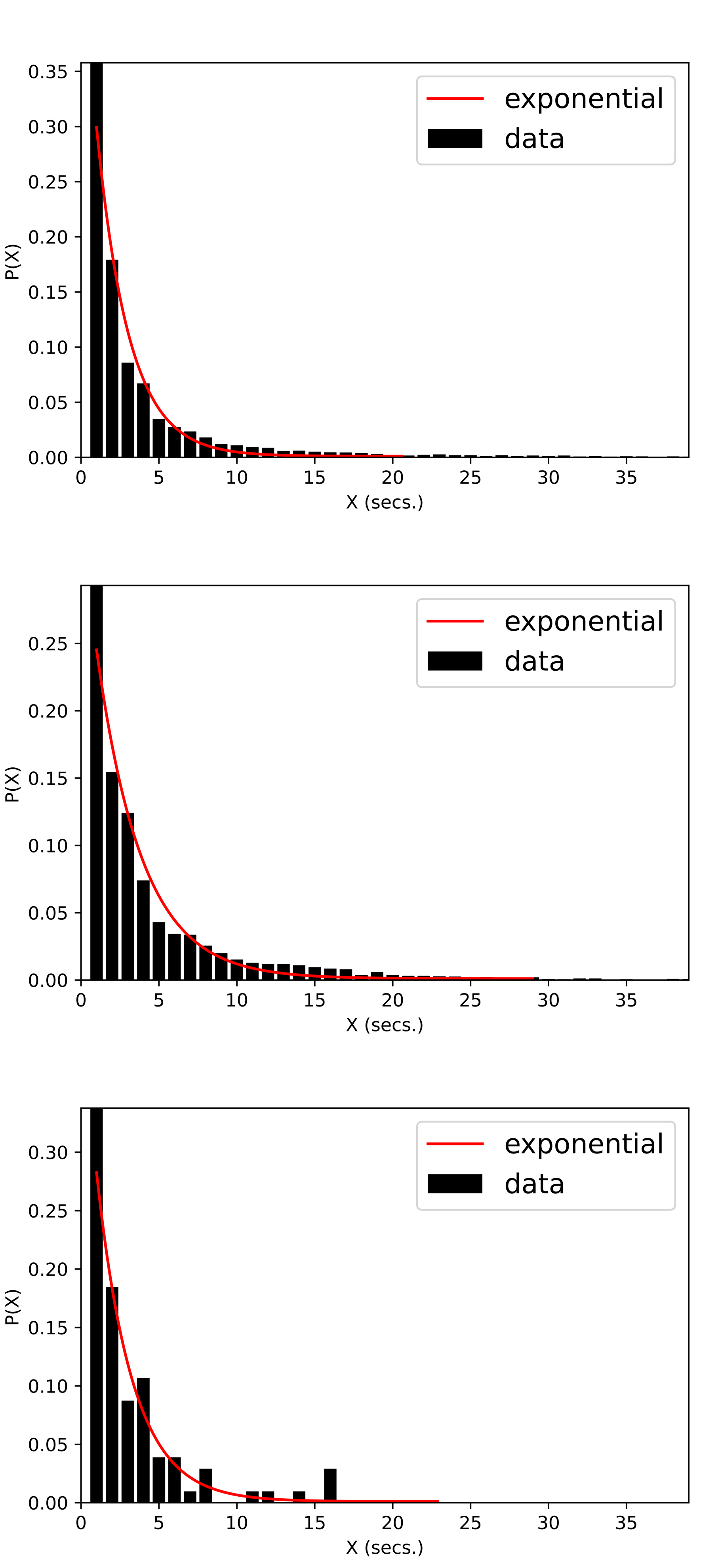}
            }
        \caption{Inactivity statistics (Older Adult Participants 1-3).}
        \label{fig.13}
        \end{figure}

Table \ref{tab7} presents the overall percentages of inactivity periods in different time ranges among all participants, showing that more than 90\% of the inactivity instances were under 10 seconds, and approximately 99\% of the instances were under 30 seconds.
On the other hand, several periods of inactivity over 100 seconds and one over 800 seconds were observed from Participant 1, who was napping on the couch at the time.
Fig. \ref{fig.11} shows an example of monitoring data for Participant 4, where the participant’s presence was between 10 am and 22 pm, with a peak inactivity duration at 19 pm. 
Meanwhile, the exponential distribution fits all the inactivity data well (see also Fig. \ref{fig.13} for Participants 1-3). 

{The authors acknowledge that one cannot make strong claims on the basis of only 4 participants, but it is hoped that the similarity of the inactivity distributions in Figures \ref{fig.10}, \ref{fig.11}, and \ref{fig.13} across the four volunteers suggests that it is possible to discriminate between normal short periods of inactivity and more serious longer periods.}

\begin{table}[h]
\centering
\caption{Average Duration (seconds) of Inactivity for Four Older Adult Participants}
\label{tab6}
\begin{tabular}{cccc}
   \toprule
  & {Median} & {Max25\%} & {Min25\%} \\
  \midrule
P1 & 2.07                       & 15.61                       & 1.37                        \\
P2 & 2.90                       & 13.76                       & 1.26                        \\
P3 & 2.30                       & 7.58                        & 1.43                        \\
P4 & 2.40                       & 9.31                        & 1.12    \\
Mean & 2.42                      & 11.57                      & 1.30   \\
SD & 0.30                       & 3.25                        & 0.12    \\
\bottomrule
\end{tabular}
\end{table}

\begin{table*}[h]
\centering
\caption{Percentage of Inactivity Duration for Different Time Ranges}\label{tab7}%
\begin{tabular}{ccccccccc}
 \toprule
Time range (s) & [1,2) & [2,5) & [5,10) & [10,30) & [30,60) & [60,200) & [200,500) & 500+   \\
 \midrule
P1                 & 44.74\% & 33.22\% & 11.60\%  & 8.08\%    & 1.75\%    & 0.55\%     & 0.03\%      & 0.03\% \\
P2                 & 35.46\% & 35.28\% & 15.64\%  & 12.33\%   & 1.15\%    & 0.13\%     & 0.02\%      & 0.00\% \\
P3                 & 44.66\% & 37.86\% & 11.65\%  & 5.83\%    & 0.00\%    & 0.00\%     & 0.00\%      & 0.00\% \\
P4                 & 39.33\% & 38.80\% & 14.02\%  & 7.45\%    & 0.35\%    & 0.04\%     & 0.00\%      & 0.00\% \\
Mean               & 41.05\% & 36.29\% & 13.23\%  & 8.42\%    & 0.81\%    & 0.18\%     & 0.01\%      & 0.01\% \\
SD  & 3.49\% & 1.96\% & 1.52\%  & 2.15\%   & 0.61\%   & 0.20\%    & 0.01\%     & 0.01\%\\

Cumulative             & 41.05\% & 77.34\% & 90.57\%  & 98.99\%   & 99.80\%   & 99.98\%    & 99.99\%     & 100\% \\
\bottomrule
\end{tabular}
\end{table*}

\section{discussion}\label{secextra}

{In this study, the MISO was proposed, a camera-based system for monitoring inactivity among single older adults in home environments during daily activities. The zero-interaction system offers advantages over the wearable, which requires constant wearing or charging. Additionally, compared to ambient sensors, our system is multi-functional. The same device can perform various tasks using different algorithms. Furthermore, it provides high-level features that are semantically meaningful. For instance, it can understand fine-grained motion (such as which part of the body is moving) and interpret environmental context and interactions.}

{For inactivity detection, non-parametric-based methods in depth maps and motion detection methods in RGB are good for environmental adaptivity. There is no need to retrain to adapt to new environments. 
The proposed method is fast enough to be used in a compact processor in real-time as well. 
The system was tested in different scenarios, and TV conditions at different distances, and the results remain constant in indoor scenarios for motion detection.
The results show that compared to SOTA pre-trained models, our method excelled in accurately detecting small body motion, demonstrated robustness in low-light conditions, as well as resistance to environmental factors such as TV light flickering and the presence of pets.}

{The recorded anonymized data can reveal the activity characteristics of older adults, such as daily habits of body movement patterns while staying at their favorite home places. 
It offers real-time tracking, enabling timely detection of excessive inactivity events. 
Additionally, because the approach saves longer-term inactivity statistics, it supports the analysis of chronic mobility issues.}

{Currently, the real-time text data is captured and stored locally for privacy-preserving, without the need for an internet connection. This facilitates long-term mobility records, such as weekly or monthly inactivity distributions. The system can also be configured to provide local user reminders, for example, a blinking light on the device to encourage older adults to stay active. However, for emergency situations involving critical inactivity events like falls or loss of consciousness, future internet connectivity would be necessary to enable identification and localization of the user. This integration would necessitate careful consideration of infrastructure safety and privacy concerns.}

{The final decisions regarding how the monitoring data is used will depend on factors outside the scope of this paper, such as the target user's behavior context, medical conditions, stakeholders, and relevant healthcare policies.}

\section{Conclusion}\label{sec6}
A system for inactivity detection in older adult residents' homes using an RGB-D camera and a small computer processor was presented.
Collecting several days of data from each local household characterized the device's performance under {real-home} conditions. 
The method was tested in different living environments and various lighting conditions.
Data processing for analysis is carried out in real-time. The system runs the inactivity detection task at 3--5 frames per second. 
The devices are small, data is anonymous, unobtrusive, and low-cost, which can be distributed well at homes for long-term use. 
A lack of motion is unlikely to be missed (a 0\% false positive rate with $\pm$3 frames temporal tolerance), and a 3\% false negative rate of missing true body motion on controlled short-term experiments. True lack of motion is likely to be long-term and this will be noticed. 

The main limitation of the described method is its performance in extremely dark environments, where the object detector often fails to detect a person or a pet. Additionally, the study faces constraints related to the small dataset and the lack of ground truth for real-life data.

Future work should focus on enhancing detection methods, capturing long-term personalized behavioral profiles of individual older adults to identify slowly deteriorating conditions, and designing decision rules based on historical patterns to provide warnings about potentially dangerous medical situations.

\section{Acknowledgments}

This research was funded by the Legal \& General Group (research grant to establish the independent Advanced Care Research Centre at the University of Edinburgh). The funder had no role in the conduct of the study, interpretation or the decision to submit for publication. The views expressed are those of the authors and not necessarily those of Legal \& General. 
{Approval for the experiments was granted by the School of Informatics Ethics Committee.}

\bibliographystyle{ACM-Reference-Format}
\bibliography{Monitoring}

\end{document}